\documentclass[conference]{IEEEtran}
\usepackage{cite}
\usepackage{graphicx}
\graphicspath{ {images/} }

\usepackage{subfig}
\usepackage{booktabs}
\usepackage{amsfonts}
\usepackage{amsmath}

\begin{document}

\pdfinfo{/Title (Unsupervised Learning with Self-Organizing Spiking Neural Networks) /Author (Hananel Hazan, Daniel Saunders, Darpan T. Sanghavi, Hava Siegelmann, Robert Kozma)}

\title{Unsupervised Learning with Self-\\Organizing Spiking Neural Networks}

\author
{\IEEEauthorblockN{Hananel Hazan\IEEEauthorrefmark{1}, Daniel Saunders\IEEEauthorrefmark{2}, Darpan T. Sanghavi\IEEEauthorrefmark{3}, \\Hava Siegelmann\IEEEauthorrefmark{4}, Robert Kozma\IEEEauthorrefmark{5}} 
\IEEEauthorblockA{
College of Information and Computer Sciences \\
University of Massachusetts Amherst \\
140 Governors Drive \\
Amherst, MA 01003, USA \\
Email: \{hhazan\IEEEauthorrefmark{1}, djsaunde\IEEEauthorrefmark{2}, dsanghavi\IEEEauthorrefmark{3}, hava\IEEEauthorrefmark{4}, rkozma\IEEEauthorrefmark{5}\}@cs.umass.edu
}
}
\maketitle

\begin{abstract}

We present a system comprising a hybridization of self-organized map (SOM) properties with spiking neural networks (SNNs) that retain many of the features of SOMs. Networks are trained in an unsupervised manner to learn a self-organized lattice of filters via excitatory-inhibitory interactions among populations of neurons. We develop and test various inhibition strategies, such as growing with inter-neuron distance and two distinct levels of inhibition. The quality of the unsupervised learning algorithm is evaluated using examples with known labels. Several biologically-inspired classification tools are proposed and compared, including population-level confidence rating, and n-grams using spike motif algorithm. Using the optimal choice of parameters, our approach produces improvements over state-of-art spiking neural networks.

% Although these networks do not reach state of the art classification performance, they achieve good accuracy with only a small number of examples and are robust to incomplete connectivity. The degree of lattice smoothness is varied and its effect on the classification performance is analyzed. The MNIST dataset is used as the experimental testbed and improvements are demonstrated over the performance and convergence speed of a baseline SNN.

\end{abstract}

\section{Introduction}
\label{intro}

Powerful deep learning approaches that dominate AI today use ``global'' gradient-based learning \cite{werbos74, pdp86}. The success of deep learning \cite{y._lecun_deep_2015} is based on using massive amounts of data to train the networks, which requires significant computational resources. In some practical problems, however, we may not have a sufficiently large data set to span the problem space, or we may need to make a decision fast, without an expensive training process. There are several approaches to circumvent the above mentioned constraints of deep learning. Some of these alternatives are based on local learning rules, such as the biologically motivated Spike-Timing Dependent Plasticity (STDP) \cite{bi_synaptic_1998}. These approaches address a trade-off between the inevitably decreased classification performance due to the unsupervised nature of the model, and the advantage provided by the distributed nature of the local training architecture. Substituting global learning with suitable local learning rules can provide a solution to the computational bottleneck of deep learning, by striking a balance between significantly increased learning speed and less requirements for computational memory resources at the cost of slightly reduced accuracy of learning and classification.

In order to run the training phase of the spiking network models, one records synaptic traces (for modifiable synapses) and conductances and neuron membrane potentials. Certain spiking neural networks have the potential advantage of being more memory efficient than their deep learning counterparts. Deep neural networks must cache the results of its layer-wise computation produced during its forward pass, and use these cached results to compute its backward pass. This effect is proportional to the number of layers utilized in the network. Studying the convergence of the training shows that the number of data samples needed to train the model to a reasonable level of accuracy was much fewer than in traditional networks.

There is extensive literature of spiking neural networks in various applications \cite{w._gerstner_spiking_2002}. Depending on the requirements of the SNN simulations, various packages provide many different levels of biological plausibility. For example, the works by \cite{Gewaltig:NEST, d._f._m._goodman_brian_2009} provide high biological realism, while \cite{CARLsim} focus more on computational efficiency.

%Deep learning has demonstrated a remarkable ability to classify and cluster data. These abilities come with great cost, both on computational resources and data requirements \cite{y._lecun_deep_2015}. Without large labeled data and lack of computational resources for training, however, the deep learning approach may fail. Although unsupervised deep learning can perform remarkably, further development in spiking neural networks trained via unsupervised learning algorithms may give biologically-inspired motivation to the development of more general and flexible artificial intelligence.

%Motivated by the shortcomings of deep learning and inspired by brain computation, we are developing an unsupervised learning framework using spiking neural networks in lieu of the popular deep learning approach. Interestingly, SNN training and prediction has the potential to be \textit{massively parallelized} while remaining \textit{energy efficient}, and scale modestly with the size of input data and number of neurons used for computation. 

%Building on the success of the SNNs studied in \cite{p._u._diehl_unsupervised_2015}, several modifications are introduced to improve performance, incorporate several SOM properties, and improve the speed at which networks learn. We change connectivity patterns between and within neuron populations to achieve an unsupervised clustering.

An interesting implementation of spiking neural networks for image processing using the MNIST handwritten digit database was published by \cite{p._u._diehl_unsupervised_2015}, and further studied by \cite{saundersetal18}. This implementation uses a modified version of the leaky integrate-and-fire (LIF) spiking neuron \cite{w._gerstner_spiking_2002} on the Brian platform \cite{d._f._m._goodman_brian_2009}. For details of the results, see \cite{saundersetal18}.

We extend the work presented in \cite{saundersetal18} by combining STDP rules with a version of the Self-Organizing Map (SOM) algorithm \cite{kohonen_self-organizing_2001}. One of the properties of SOMs is the ability to cluster an unlabeled dataset in an unsupervised manner. The topological arrangement created by the SOM algorithm forms clusters that specialize and are unique to categories that exist in the input data. This clustering property is reminiscent of specialized areas in the primate cortex, where groups of neurons organize themselves according to similar functionality of parts of the primate body \cite{doi:10.1162/neco.1995.7.3.425}. This property has never been addressed in the SNN literature. This paper aims to amend this issue. We show examples of SNN learned representations, and present classification results based on network filters and spiking activity on the MNIST handwritten digit dataset \cite{y._lecun_gradient-based_1998}.

\section{Methods}
\label{Methods}

\subsection{Network architecture}

The spiking neural network has a multi-layer architecture, including input layer, excitatory layer, and inhibitory layer. The input layer is an array of $N \times N$ neurons, which correspond to the pixel of the input images in an image processing application. The excitatory and inhibitory layers have size of $K \times K$. Each input pixel projects to each of the excitatory nodes through modifiable synapses. Spikes are generated in the input neurons using Poisson spiking neurons, with average firing rate $\lambda$ proportional to pixel intensity \cite{w._gerstner_spiking_2002}. This process is simulated for a specific time duration, and the STDP rule is executed for the connection between the winning neuron in the excitatory layer and the corresponding input pixel. It is an important property of the model that once a neuron fires, it inhibits all other neurons according to the inhibition scheme described below. For further details of the basic model structure, including considerations on the system size, see \cite{saundersetal18}.

%This behavior is what allows individual neurons to learn unique filters. Increasing the number of excitatory and inhibitory neurons has the effect of allowing a SNN to \textit{memorize} more data examples from the training data and recognize similar patterns during the test phase. This network architecture is illustrated in Figure \ref{fig:Spiking neural network architecture}, and an example set of filters (for an SNN with 400 excitatory and inhibitory neurons) is shown in Figure \ref{fig:Learned filters}. Although we arrange the filters in a two-dimensional matrix, there is no neighborhood structure between the learned filters; i.e., the excitatory neurons do not communicate their firing activity to their neighbors.

\begin{figure}
  \centering
  \includegraphics[width=0.45\textwidth, height=5.75cm]{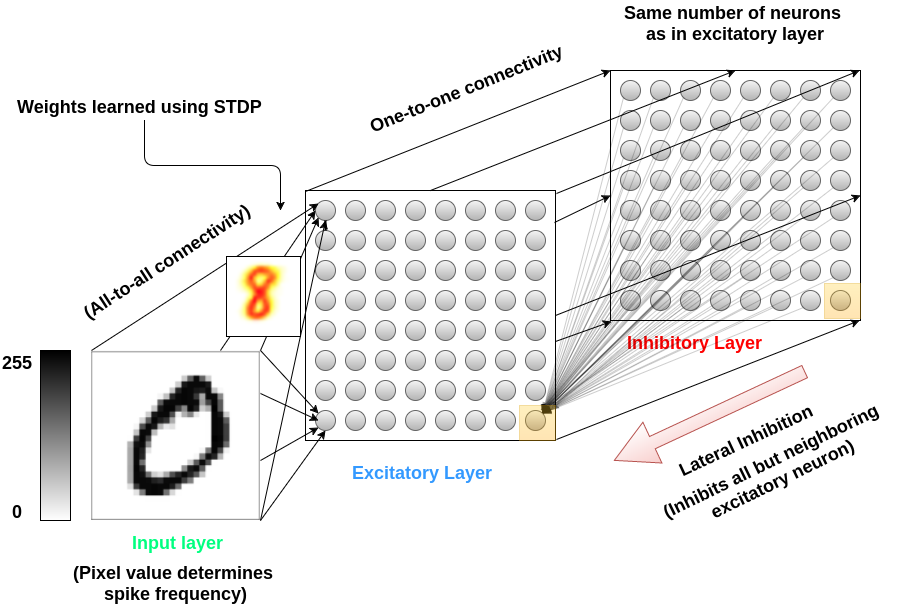}
  \caption{Spiking neural network architecture; from \cite{saundersetal18}.}
  \label{fig:Spiking neural network architecture}
\end{figure}

\subsection{Increasing inhibition with inter-neuron distance}

We make a change to the SNN architecture which is inspired by the self-organizing map (SOM) and corresponding algorithm \cite{kohonen_self-organizing_2001}. This change is included in part to curb the degree of \textit{competition} imposed by the connections from the inhibitory layer introduced in \cite{p._u._diehl_unsupervised_2015}, in which network activation is too sparse throughout training to learn filters quickly. Additionally, this change causes network filters to self-organize into distinct clusters by classes of data.

\begin{figure} 
  \centering
  \includegraphics[width=0.45\textwidth, height=7cm]{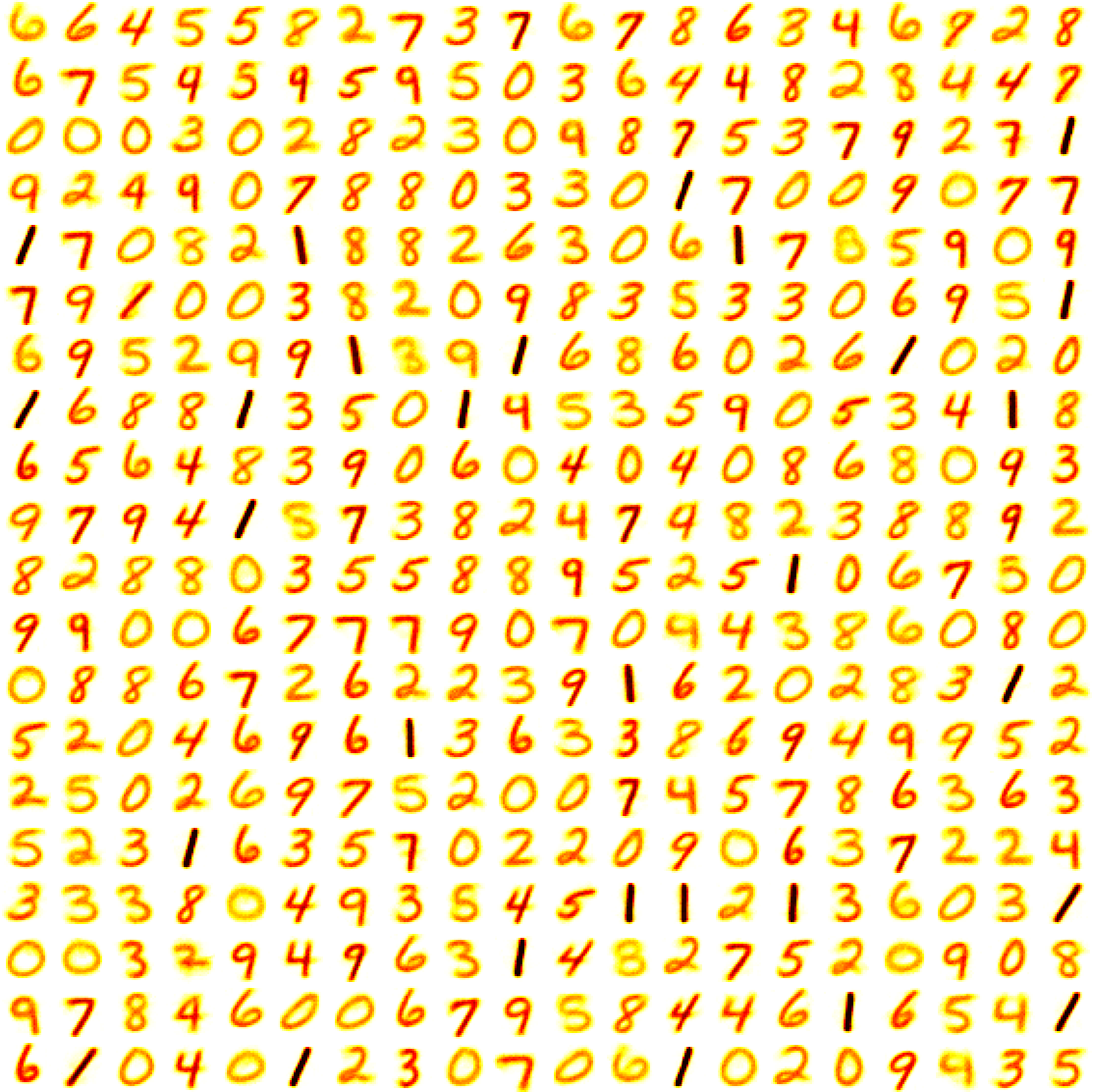}
  \caption{Learned filters from baseline model}
  \label{fig:Learned filters}
\end{figure}

\begin{figure}[!tbp]
  \centering
  \includegraphics[width=0.475\textwidth]{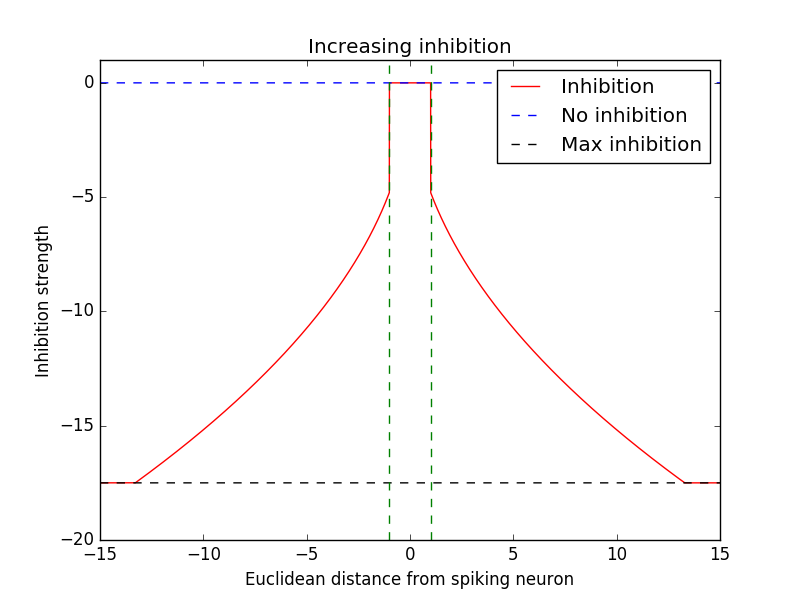}
  \caption{Inhibition as a function of Euclidean distance}
  \label{fig:Inhibition as a func}
\end{figure}

Instead of inhibiting all other neurons at a large fixed rate as in \cite{saundersetal18, p._u._diehl_unsupervised_2015}, we increase the level of inhibition with the distance between neurons. Inhibition level is increased in proportion to the square root of the Euclidean distance, similar to the SOM learning algorithm. This requires a parameter $c_{\textrm{inhib}}$ which is multiplied by the distance to compute inhibition levels, as well as a parameter $c_{\textrm{max}}$ that gives the maximum allowed inhibition (see Figure \ref{fig:Inhibition as a func}). With proper choice of $c_{\textrm{inhib}}$, when a neuron exceeds its firing threshold $v_{\textrm{threshold}}$, instead of inhibiting all other neurons from firing for the rest of the iteration, a neighborhood of nearby neurons will be weakly inhibited and may have the chance to fire. This encourages neighborhoods of neurons to fire for the same inputs and learn similar filters. See Figure \ref{fig:Two-level figure} for plots of the effect of increasing inhibition to create clusters of filters. Compare this to Figure \ref{fig:Learned filters}, in which filters are learned by using fixed inhibition with the learning algorithm described in \cite{p._u._diehl_unsupervised_2015}.

To avoid confusion, we call SNNs with all variants of the \textit{increasing inhibition} strategy \textit{lattice map spiking neural networks}, abbreviated as LM-SNNs.

\begin{figure}[!tbp]
  \centering
  \subfloat[25x25 lattice of neuron filters]{
  \includegraphics[width=0.215\textwidth]{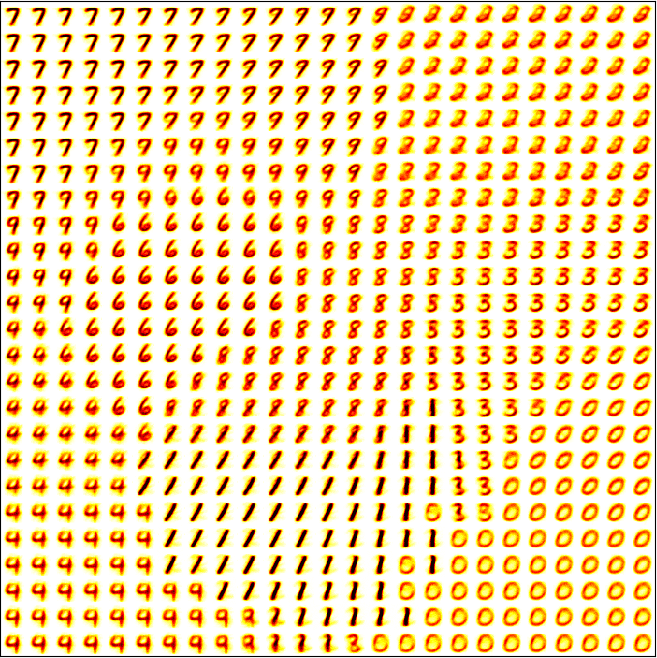}
  \label{fig:increasing_filters}}
  \subfloat[Neuron class assignments]{
  \includegraphics[width=0.240\textwidth]{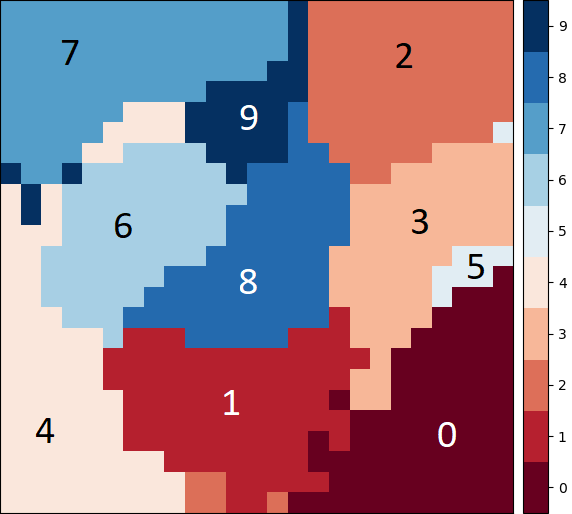}
  \label{fig:increasing_assignments}}
  \caption{\textit{Increasing inhibition}: Filter map and class assignments}
  \label{fig:increasing figure}
\end{figure}

\subsection{Growing the inhibition level over the training phase}

We want to produce \textit{individualized} filters as learned by the SNN presented in \cite{p._u._diehl_unsupervised_2015}, yet retain the clustering of filters achieved by our \textit{increasing inhibition} modification. Distinct filters are necessary to ensure that our learned representation contains as little redundancy as possible, making the best use of model capacity. To that end, we implemented another modification to the inhibition scheme, where the inhibition constant $c_{\textrm{inhib}}$ grows on a linear schedule from a small value $c_{\textrm{min}} \approx 0.5$ to a large value $c_{\textrm{max}} \approx 17.5$. The \textit{increasing inhibition} strategy is used as before; however, by the end of network training, the inhibition level is equivalent to that of \cite{p._u._diehl_unsupervised_2015}. In this setting, the filters self-organize into smoothly-varying clusters, and then individualize as the inhibition level becomes large. We also consider growing the inhibition level to $c_{\textrm{max}}$ for some percentage of the training ($p_{\textrm{grow}}$) and holding it fixed for the rest ($1 - p_{\textrm{grow}}$). During the last $1 - p_{\textrm{grow}}$ proportion of the training phase, neuron filters are allowed to individualize more, allowing them to remember less frequent prototypical examples from the data.

\subsection{Two-level inhibition}

To remove the need to re-compute inhibitory synapse strengths throughout network training, we implemented a simple \textit{two-level inhibition} scheme: For the first $p_{\textrm{low}}$ proportion of the training, the network is trained inhibition level $c_{\textrm{min}}$; for the last $1 - p_{\textrm{low}}$ proportion, the network is trained with $c_{\textrm{max}}$. The inhibition level is not smoothly varied between the two levels, but jumps suddenly at the $p_{\textrm{low}}$ mark. At the low inhibition level, large neighborhoods of excitatory neurons compete \textit{together} to fire for certain types of inputs, eventually organizing their filters into a SOM-like representation of the dataset. At the high inhibition level, however, neurons typically maintain yet sharpen the filter acquired during the low inhibition portion of training. In this way, we obtain filter maps similar to those learned using the \textit{growing inhibition} mechanism, but with a simpler implementation. This inhibition strategy represents a middle ground between that of \cite{p._u._diehl_unsupervised_2015} and our \textit{increasing inhibition} scheme.

\begin{figure}[!tbp]
  \subfloat[25x25 lattice of neuron filters]{
  \includegraphics[width=0.21\textwidth,height=5cm,keepaspectratio]{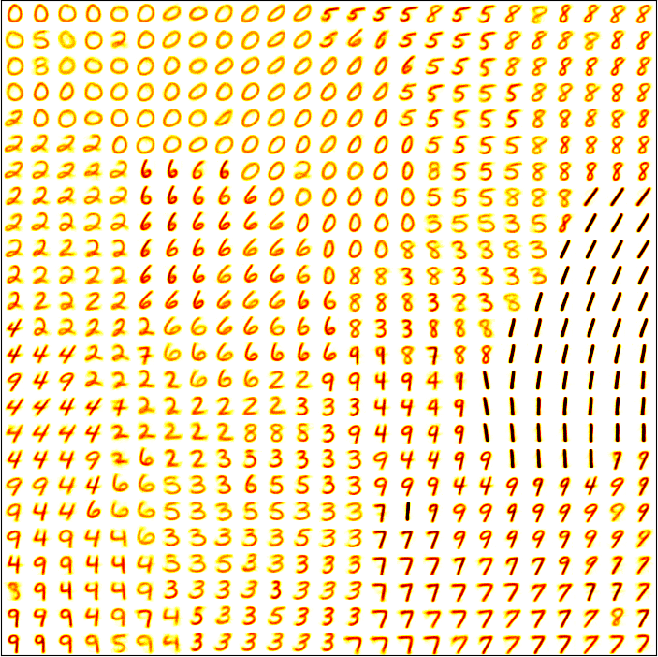}
  \label{fig:two_level_filters}}
  \subfloat[Neuron class assignments]{
  \includegraphics[width=0.239\textwidth,height=5cm,keepaspectratio]{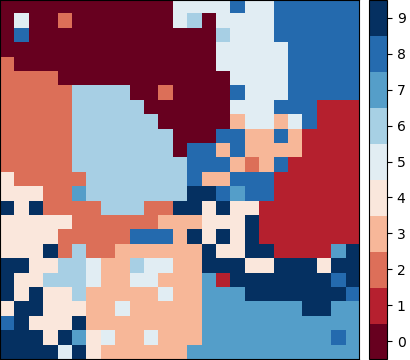}
  \label{fig:two_level_assignments}}
  \caption{\textit{Two-level inhibition}: Filter map and class assignments}
  \label{fig:Two-level figure}
\end{figure}

See Figure \ref{fig:Two-level figure} for an example learned filter map and neuron class assignments. There is some degree of clustering of the filters; however, as the inhibition level approaches that of \cite{p._u._diehl_unsupervised_2015}, they may eventually move away from the digit originally representing on their weights, \textit{fragmenting} the filter clustering. The degree of this fragmentation depends on the choice of $p_{\textrm{low}}$: with more time training with the $c_{\textrm{max}}$ inhibition level, the more likely a neuron is to change its filter to represent data outside of its originally converged class.

\subsection{Evaluating learned representations}

Although LM-SNNs are trained in an unsupervised manner, we may want to evaluate the quality of the representations they learn. The dataset representation is encoded in the learned weights of synapses connecting the input and excitatory layers. We use the activity of the neurons in the excitatory layer with their filter weights held fixed to (1) say what it means to \textit{represent} input data from a certain class, and (2) \textit{classify} new inputs based on historical network activity.

We perform a two-step procedure before the test phase to \textit{label} excitatory neurons with the input category we believe they represent, and then \textit{classify} new data based on these labels and the spiking activity of the excitatory neurons on it, as in \cite{p._u._diehl_unsupervised_2015}. We call the first step \textit{neuron labeling} and the second \textit{voting}. For some \textit{voting schemes}, neurons are assigned the label of the input class for which they have fired most on average, and these labels are used to classify new data based on network activation. In the \textit{all} voting scheme, all excitatory neuron spikes are counted in the ``vote'' for the label of the input example. In the \textit{confidence weighting} voting scheme, we record the proportions of spikes each neuron fires for each input class, and used a weighted sum of these proportions and network activity to classify new inputs. These evaluation strategies are reminiscent of related work \cite{Gollisch1108} in which the timing of individual neurons' spikes are discarded in favor of a rate-based code.

In order to leverage the information contained in the timing of the spikes, we also design an n-gram based testing scheme that considers the order of spiking to make a prediction. N-grams have long been used for sequential modeling, particularly in computational linguistics. Our \textit{n-gram} scheme, like the rate-based schemes, also follows a two-step procedure: a \textit{learning} phase to estimate the n-gram conditioned class probabilities from a subset of the training data, and a \textit{voting} phase where the n-grams in the output spiking sequence ``vote'' for the classes. While this scheme does not exploit the spike timing explicitly, our motivation for now is only to demonstrate the importance of the information contained in spike ordering. There is growing evidence \cite{PETERSEN2001503} on how most of the information in cortical neuronal networks is contained in the timing of individual spikes, and in particular on the crucial importance of the exact time of the first spike. Moreover, n-grams are able to identify repeating motifs \cite{RAICHMAN200896} in the activation of synchronized bursts in cultured neuronal networks and can also classify stimuli based on the time to first spike \cite{Kermany9588}.

The \textit{distance} voting scheme is used to benchmark the performance of the spike-based schemes: new data is labeled with the class label of the neuron whose filter most closely matches the input data as measured by Euclidean distance.

In Section \ref{sec:Results}, we evaluate networks with the \textit{all}, \textit{confidence weighting}, \textit{distance} and \textit{n-gram} voting schemes. Other voting schemes were designed and tested, but eventually discarded in favor of those which produced the most consistent accuracy results.

\section{Results}
\label{sec:Results}

% \subsection{Linearly Growing Inhibition}
% \label{ssec:growing inhibition}

% Networks are trained on a single pass through the MNIST training data, comprising 60K examples, and evaluated on all 10K test examples. Test accuracy results for networks of size 400, 625, and 900 are shown in Table \ref{tab:growing_accuracy}, each averaged over 10 independent trials (\textbf{TODO}).

% \begin{table}
% \begin{center}
%  \caption{Growing inhibition test accuracy}
%  \label{tab:growing_accuracy}
%  \begin{tabular}{||c|c|c||}
%  \hline
%  $n_e$, $n_i$ & $p_{\textrm{grow}}$ & Test accuracy \\ [0.5ex] 
%  \hline
%  400 & 25\% & \textbf{91.48}\% \\
%  400 & 50\% & 91.32\% \\ 
%  400 & 75\% & 89.63\% \\ 
%  400 & 100\% & 89.83\% \\ 
%  \hline
%  625 & 25\% & \textbf{91.71}\% \\
%  625 & 50\% & 91.41\% \\ 
%  625 & 75\% & 91.51\% \\ 
%  625 & 100\% & 90.63\% \\ 
%  \hline
%  900 & 25\% & (seizure) \\
%  900 & 50\% & \textbf{93.06}\% \\ 
%  900 & 75\% & 92.95\% \\ 
%  900 & 100\% & 92.53\% \\ 
%  \hline
% \end{tabular}
% \end{center}
% \end{table}

In the subsequent sections, we give quantitative results of variants of the LM-SNN models. We omit results on the \textit{increasing} and \textit{growing inhibition} strategies in favor of results using the simpler \textit{two-level inhibition} strategy in Section \ref{ssec:two-level inhibition}. A large compilation of results demonstrates that network performance is robust to a wide range of hyper-parameter choices. This strategy is compared with a baseline SNN of \cite{p._u._diehl_unsupervised_2015} in Section \ref{ssec:comparing SNN and two-level}, and is shown to outperform it, especially in the regime of small networks, and particularly so with the \textit{n-gram} voting scheme. Section \ref{ssec:Training larger networks} shows multiple passes through training data are required for successful training of progressively larger networks. An improved rate of convergence to near-optimal performance is quantitatively demonstrated in Section \ref{ssec:convergence}, using baseline SNNs of \cite{p._u._diehl_unsupervised_2015} to compare. Finally, in Section \ref{ssec:sparse} we discuss the robustness of LM-SNNs in the absence of inputs, showing a graceful degradation in performance.

\subsection{Two-level inhibition}
\label{ssec:two-level inhibition}

Networks are trained on a single pass over the training data and evaluated on all 10K test examples. Test accuracy results and single standard deviations are reported in Table \ref{tab:two_level_accuracy}, each averaged over 10 independent trials. All networks are comprised of 625 excitatory and inhibitory neurons. Results are demonstrated for a number of choices of parameters $p_{\textrm{low}}$, $c_{\textrm{min}}$, and $c_{\textrm{max}}$.

\begin{table*}
\begin{center}
 \caption{Two-level inhibition test accuracy ($n_e, n_i$ = 625)}
 \label{tab:two_level_accuracy}

\begin{tabular}{||c|c|c|c|c|c||c|c|c|c|c|c||}
\toprule
$p_{\textrm{low}}$ & $c_{\textrm{min}}$ & $c_{\textrm{max}}$ & \textit{distance} &      \textit{all} & \textit{confidence} & $p_{\textrm{low}}$ & $c_{\textrm{min}}$ & $c_{\textrm{max}}$ & \textit{distance} &      \textit{all} & \textit{confidence} \\
\midrule
               0.1 &                0.1 &               15.0 &   91.8 $\pm$ 0.63 &   91.4 $\pm$ 0.13 &    91.69 $\pm$ 0.63 &			  0.25 &                0.1 &               15.0 &  91.51 $\pm$ 0.25 &  90.62 $\pm$ 0.26 &    90.97 $\pm$ 0.25 \\
               0.1 &                0.1 &               17.5 &  92.02 $\pm$ 0.38 &  91.26 $\pm$ 0.11 &    91.68 $\pm$ 0.38 &              0.25 &                0.1 &               17.5 &  91.83 $\pm$ 0.18 &  91.06 $\pm$ 0.18 &    91.54 $\pm$ 0.18 \\
               0.1 &                0.1 &               20.0 &  92.38 $\pm$ 0.49 &  91.54 $\pm$ 0.14 &     92.1 $\pm$ 0.49 &              0.25 &                0.1 &               20.0 &  92.16 $\pm$ 0.34 &  91.07 $\pm$ 0.29 &    91.83 $\pm$ 0.34 \\
               0.1 &                1.0 &               15.0 &  91.67 $\pm$ 0.63 &  91.12 $\pm$ 0.39 &    91.59 $\pm$ 0.63 &              0.25 &                1.0 &               15.0 &  91.36 $\pm$ 0.24 &  90.26 $\pm$ 0.21 &    90.77 $\pm$ 0.24 \\
               0.1 &                1.0 &               17.5 &  92.25 $\pm$ 0.42 &  91.32 $\pm$ 0.29 &    91.67 $\pm$ 0.42 &              0.25 &                1.0 &               17.5 &  91.78 $\pm$ 0.61 &  91.09 $\pm$ 0.29 &    91.46 $\pm$ 0.61 \\
               0.1 &                1.0 &               20.0 &  92.36 $\pm$ 0.66 &  91.44 $\pm$ 0.38 &     91.9 $\pm$ 0.66 &              0.25 &                1.0 &               20.0 &   \textbf{92.3 $\pm$ 0.19} &  91.51 $\pm$ 0.17 &    91.98 $\pm$ 0.19 \\
               0.1 &                2.5 &               15.0 &  91.99 $\pm$ 0.53 &  91.01 $\pm$ 0.29 &     91.3 $\pm$ 0.53 &              0.25 &                2.5 &               15.0 &  91.65 $\pm$ 0.64 &  90.89 $\pm$ 0.31 &    91.19 $\pm$ 0.64 \\
               0.1 &                2.5 &               17.5 &  92.17 $\pm$ 0.39 &  91.49 $\pm$ 0.17 &    91.86 $\pm$ 0.39 &              0.25 &                2.5 &               17.5 &  92.04 $\pm$ 0.62 &  91.52 $\pm$ 0.27 &    91.95 $\pm$ 0.62 \\
               0.1 &                2.5 &               20.0 &  \textbf{92.55 $\pm$ 0.54} &   92.07 $\pm$ 0.3 &    92.49 $\pm$ 0.54 &              0.25 &                2.5 &               20.0 &  92.26 $\pm$ 0.33 &  91.43 $\pm$ 0.48 &    91.97 $\pm$ 0.33 \\
\hline \\ [-1.0em]
               0.5 &                0.1 &               15.0 &  90.82 $\pm$ 0.28 &  90.05 $\pm$ 0.12 &    90.45 $\pm$ 0.28 &              0.75 &                0.1 &               15.0 &  90.35 $\pm$ 0.44 &  89.91 $\pm$ 0.45 &    90.42 $\pm$ 0.44 \\
               0.5 &                0.1 &               17.5 &  90.99 $\pm$ 0.44 &  90.22 $\pm$ 0.17 &    90.87 $\pm$ 0.44 &              0.75 &                0.1 &               17.5 &  90.42 $\pm$ 0.38 &  89.99 $\pm$ 0.27 &    90.48 $\pm$ 0.38 \\
               0.5 &                0.1 &               20.0 &  \textbf{91.78 $\pm$ 0.16} &  90.85 $\pm$ 0.32 &    91.42 $\pm$ 0.16 &              0.75 &                0.1 &               20.0 &  90.86 $\pm$ 0.31 &  90.31 $\pm$ 0.18 &    90.91 $\pm$ 0.31 \\
               0.5 &                1.0 &               15.0 &  91.09 $\pm$ 0.22 &  89.92 $\pm$ 0.19 &    90.49 $\pm$ 0.22 &              0.75 &                1.0 &               15.0 &  90.41 $\pm$ 0.11 &  89.22 $\pm$ 0.24 &    89.98 $\pm$ 0.11 \\
               0.5 &                1.0 &               17.5 &  91.18 $\pm$ 0.16 &  90.39 $\pm$ 0.46 &    90.97 $\pm$ 0.16 &              0.75 &                1.0 &               17.5 &   90.57 $\pm$ 0.3 &  89.49 $\pm$ 0.42 &     90.07 $\pm$ 0.3 \\
               0.5 &                1.0 &               20.0 &  91.57 $\pm$ 0.32 &  90.88 $\pm$ 0.79 &    91.35 $\pm$ 0.32 &              0.75 &                1.0 &               20.0 &  91.01 $\pm$ 0.16 &   89.9 $\pm$ 0.31 &    90.44 $\pm$ 0.16 \\
               0.5 &                2.5 &               15.0 &   91.3 $\pm$ 0.38 &  90.45 $\pm$ 0.35 &    90.95 $\pm$ 0.38 &              0.75 &                2.5 &               15.0 &  90.68 $\pm$ 0.27 &   89.5 $\pm$ 0.23 &     90.1 $\pm$ 0.27 \\
               0.5 &                2.5 &               17.5 &  91.45 $\pm$ 0.23 &  90.71 $\pm$ 0.37 &    91.24 $\pm$ 0.23 &              0.75 &                2.5 &               17.5 &  90.82 $\pm$ 0.38 &  89.76 $\pm$ 0.34 &     90.4 $\pm$ 0.38 \\
               0.5 &                2.5 &               20.0 &  91.72 $\pm$ 0.27 &  91.04 $\pm$ 0.38 &    91.63 $\pm$ 0.27 &              0.75 &                2.5 &               20.0 &   \textbf{91.21 $\pm$ 0.2} &  90.47 $\pm$ 0.51 &     91.01 $\pm$ 0.2 \\
\bottomrule
\end{tabular}

\end{center}
\end{table*}

\subsection{Comparing baseline SNN and two-level inhibition}
\label{ssec:comparing SNN and two-level}

To compare how the two-level inhibition strategy performs with respect to the networks of \cite{p._u._diehl_unsupervised_2015}, we present in Table \ref{tab:eth_vs_two_level} a comparison of their accuracies over several settings of the number of neurons, using the \textit{confidence}, \textit{all}, \textit{distance} and \textit{n-gram} voting schemes. All networks are trained for 60K iterations (a single pass through the training data) and evaluated on the 10K examples from the test data. The two-level inhibition hyper-parameters are fixed to $p_{\textrm{low}} = 0.1, c_{\textrm{min}} = 0.1, c_{\textrm{max}} = 20.0$. For the \textit{n-gram} scheme, 12K examples are used for the learning phase. Preliminary tests showed that bi-grams gave the best performance, and hence we fixed $n = 2$. 5 independent experiments with different initial configurations and Poisson spike trains were run, and their results are averaged and reported along with a single standard deviation. 

One can notice the superiority of the \textit{confidence} scheme to the \textit{all} scheme, the \textit{distance} scheme to the \textit{confidence} scheme, and the \textit{n-gram} scheme to the \textit{distance} scheme. While the \textit{confidence}, \textit{all} and \textit{n-gram} schemes use the activity of the network in order to classify new data, the \textit{distance} scheme simply labels new inputs with the label of the neuron whose filter most closely matches the input. This last evaluation scheme is reminiscent of the one-nearest neighbor algorithm. However, our spiking neural networks learn prototypical data vectors, whereas the one-nearest neighbor method stores the entire dataset to use during evaluation.

% The \textit{n-gram} scheme uses the ordering of excitatory layer spikes on the training data to classify new examples.
 
\begin{table*}[ht]
 \begin{center}
 \caption{baseline SNN vs. Two-Level Inhibition SNN (60K train / 10K test)}
 \label{tab:eth_vs_two_level}
 
 \begin{tabular}{||c|c|c|c|c|c|c||}
 \hline
 $n_e$, $n_i$ & Baseline SNN & Two-level (\textit{confidence}) & Two-level (\textit{all}) & Two-level (\textit{distance}) & Two-level (\textit{n-gram}) \\ [0.5ex]
 \hline \\ [-1.0em]
 100 & 80.71\% $\pm$ 1.66\% & 82.94\% $\pm$ 1.47\% & 81.12\% $\pm$ 1.96\% & 85.11\% $\pm$ 0.74\% & 85.71\% $\pm$ 0.85\% \\
 225 & 85.25\% $\pm$ 1.48\% & 88.49\% $\pm$ 0.48\% & 87.33\% $\pm$ 0.59\% & 89.11\% $\pm$ 0.37\% & 90.50\% $\pm$ 0.43\% \\
 400 & 88.74\% $\pm$ 0.38\% & 91\% $\pm$ 0.56\% & 90.56\% $\pm$ 0.67\% & 91.4\% $\pm$ 0.38\% & 92.56\% $\pm$ 0.48\% \\ 
 625 & 91.27\% $\pm$ 0.29\% & 92.14\% $\pm$ 0.50\% & 91.69\% $\pm$ 0.59\% & 92.37\% $\pm$ 0.29\% & \textbf{93.39\% $\pm$ 0.25\%} \\
 900 & 92.63\% $\pm$ 0.28\% & 92.36\% $\pm$ 0.63\% & 91.73\% $\pm$ 0.7\% & 92.77\% $\pm$ 0.26\% & 93.25\% $\pm$ 0.57\% \\
 \hline
\end{tabular}
\end{center}
\end{table*}

\subsection{Training larger networks}
\label{ssec:Training larger networks}

One can notice in Table \ref{tab:eth_vs_two_level} that networks tend to achieve better accuracy by using more neurons. However, training networks with more than 900 neurons on a single pass through the training data shows a decrease in test performance. On the other hand, training networks with multiple passes through the data preserves the upward trend in performance (see Table \ref{tab:eth_vs_two_level_larg_net}). The results for networks with 1,225 and 1,600 neurons suggest either or both of (1) the training algorithm for both of the baseline SNN and the LM-SNN does not make appropriate use of all data on a single training pass, or (2) network capacity is too large to adequately learn all filter weights with a single pass over the data. Inspection of the convergence of filter weights during training (data not shown) suggests that the training algorithm needs to be adjusted for greater data efficiency. The fact that training with more epochs improves accuracy also points to the fact that network capacity may be too high to fit with a single pass through the data.

\begin{table*}[ht]
 \begin{center}
 \caption{Large networks: baseline SNN vs. Two-Level Inhibition SNN}
 \label{tab:eth_vs_two_level_larg_net}
 
 \begin{tabular}{||c|c|c|c|c|c|c||}
 \hline
 $n_e$, $n_i$ & $n_{\textrm{train}}$ & baseline SNN & Two-level (\textit{confidence}) & Two-level (\textit{all}) & Two-level (\textit{distance}) & Two-level (\textit{n-gram}) \\ [0.5ex]
 \hline \\ [-1.0em]
 1,225 & 1 $\times$ 60K & 91.51\% $\pm$ 0.44\% & 91.38\% $\pm$ 0.89\% & 90.93\% $\pm$ 0.88\% & 92.73\% $\pm$ 0.36\% & 92.39\% $\pm$ 0.47\% \\
 1,225 & 2 $\times$ 60K & 92.37\% $\pm$ 0.22\% & 92.25\% $\pm$ 0.63\% & 91.77\% $\pm$ 0.59\% & 92.39\% $\pm$ 0.29\% & 93.69\% $\pm$ 0.32\% \\
 1,225 & 3 $\times$ 60K & 92.43\% $\pm$ 0.23\% & 92.57\% $\pm$ 0.57\% & 91.85\% $\pm$ 0.52\% & 92.48\% $\pm$ 0.29\% & 93.87\% $\pm$ 0.25\% \\
 1,600 & 1 $\times$ 60K & 90.18\% $\pm$ 0.58\% & 89.59\% $\pm$ 0.98\% & 89.26\% $\pm$ 0.94\% & 92.45\% $\pm$ 0.33\% & 92.42\% $\pm$ 0.62\% \\
 1,600 & 2 $\times$ 60K & 92.54\% $\pm$ 0.51\% & 92.61\% $\pm$ 0.51\% & 91.79\% $\pm$ 0.56\% & 92.66\% $\pm$ 0.26\% & 93.54\% $\pm$ 0.5\% \\
 1,600 & 3 $\times$ 60K & 92.80\% $\pm$ 0.49\% & 92.96\% $\pm$ 0.56\% & 92.87\% $\pm$ 0.49\% & 93.03\% $\pm$ 0.30\% & \textbf{94.07\% $\pm$ 0.46\%} \\
 \hline
\end{tabular}
\end{center}
\end{table*}

\subsection{Network convergence}
\label{ssec:convergence}

A major advantage of using a relaxed inhibition scheme is the ability to learn a reasonably good data representation while seeing only a small number of examples. In training LM-SNNs using the \textit{growing} and \textit{two-level} inhibition strategies, we observe a convergence to optimal network performance well before SNNs trained with large, constant inhibition as in \cite{p._u._diehl_unsupervised_2015}. With small inhibition, increasing with the distance between pairs of neurons, the development of filters occurs quickly, allowing the obtainment of respectable accuracy. Over the course of the training, those filters are gradually refined as the inter-neuron inhibition grows and allows the independent firing of neurons.

\begin{figure}
  \centering
  \includegraphics[width=0.475\textwidth]{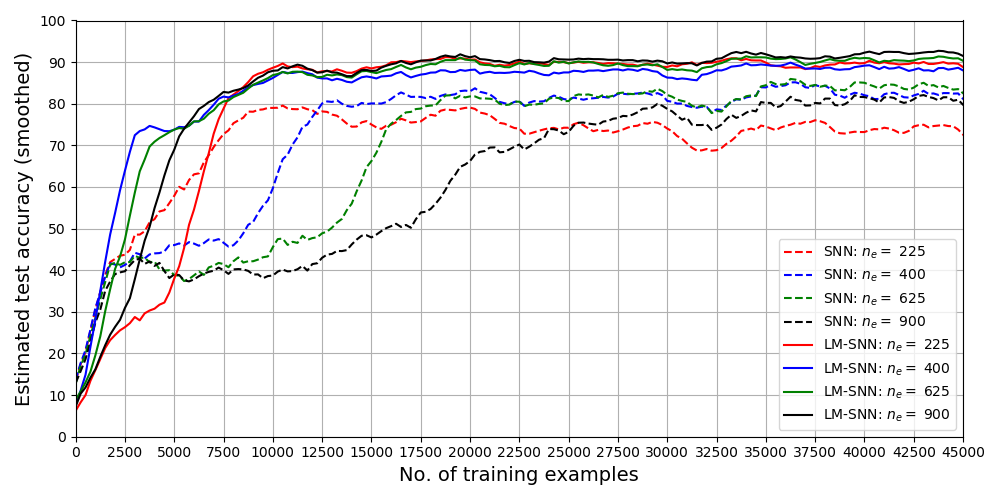}
  \caption{LM-SNN vs. SNN smoothed performance estimate over the training phase.}
  \label{fig:convergence}
\end{figure}

We show in Figure \ref{fig:convergence} a comparison of the convergence in estimated test performance for networks of various sizes. SNNs are trained with large, constant inhibition, while LM-SNNs are trained with the \textit{two-level inhibition} strategy with parameter values $p_{\textrm{low}} = 0.1$, $c_{\textrm{min}} = 1.0$, and $c_{\textrm{max}} = 20.0$. Estimates are calculated by assigning labels to neurons based on the firing activity on 250 training examples, and using these labels to classify the next 250 training examples. Training accuracy curves are \textit{smoothed} by averaging each estimate with a window of the 10 neighboring estimates. The performance of LM-SNNs quickly outpace that of SNNs, and obtain near-optimal accuracy after seeing 10 to 20 thousand training examples. Due to the inhibition strategy, larger LM-SNNs achieve better accuracy more quickly, due to their ability to learn more filters at once. On the other hand, SNNs are limited to learning one filter at a time, and their accuracy during the training phase is hindered by the size of the network as a result. 

% To compare the convergence of our LM-SNN model, we trained two-layer multi-layer perceptrons (MLPs) [citation needed?] with varying numbers of hidden units using the back-propagation algorithm \cite{rumelhart_backpropagation_1995} and the Adam optimizer [citation needed]. We also trained k-Means clustering models [citation needed] with varying numbers of clusters, and used it to classify new data by assigning labels to each cluster from the majority label of the data in each cluster, and predicting the label of the closest cluster for as-yet unseen test data examples. Both implementations used are from the \texttt{scikit-learn} machine learning library [citation needed], trained without hyper-parameter tuning.

% Smoothed training curves for these models (and LM-SNNs to compare) are shown in Figure \ref{fig:nn_kmeans_lm-snn}, where the number of hidden units (MLP) and clusters (k-Means) are selected to match the number of excitatory neurons used in the trained LM-SNNs. Interestingly, the k-Means algorithm converges to its optimal accuracy before the 2-layer MLP, and remains competitive up through 30k iterations, though the 2-layer MLP generally achieves a higher classification accuracy. Although our method converges slower than both the k-Means and MLP models, it is eventually competitive with these methods as well.

% \begin{figure}
%   \centering
%   \includegraphics[width=0.5\textwidth]{images/nn_kmeans_lm-snn.png}
%   \caption{LM-SNN vs. multi-layer perceptron (supervised) and k-Means classification (unsupervised)}
%   \label{fig:nn_kmeans_lm-snn}
% \end{figure}

\subsection{Sparse input connectivity}
\label{ssec:sparse}

Instead of connecting the input one-to-one with the layer of excitatory neurons, we experiment with varying degrees of random sparse connectivity. We are interested in whether small amounts of sparsity might make our network more robust to outliers in the MNIST data, therefore increasing the chance of good test performance. We also hope that our system will still perform well in the event of missing features, degrading in performance gracefully as the input data becomes less clear.

Interestingly, small amounts of sparsity do not degrade network performance much, and with nearly all connections removed, the network maintains reasonable accuracy. In particular, with 90\% of synapses from the input removed, networks of 625 excitatory and inhibitory neurons achieve nearly 60\% test accuracy after being trained using the \textit{two-level inhibition} mechanism. Although this technique did not result in improved test performance, it demonstrates the robustness of our system to missing information in the input data.

% Table \ref{tab:sparse_results} displays accuracy results from 10 independent training and test phases, averaged and reported along with their standard deviations. See Figure \ref{fig:sparse_filters} for a visualization of learned filters in a network with 90\% sparsity.

% \begin{table}
% \begin{center}
%  \caption{Sparse input test accuracy}
%  \label{tab:sparse_results}
%  \begin{tabular}{||c|c|c||}
%  \hline
%  $n_e$, $n_i$ & \% sparsity & Test accuracy \\ [0.5ex] 
%  \hline
%  625 & 0\% & \textbf{91.71}\% $\pm$ 0.23\% \\
%  625 & 10\% & 91.48\% $\pm$ 0.31\% \\
%  625 & 25\% & 89.79\% $\pm$ 0.66\% \\
%  625 & 50\% & 85.83\% $\pm$ 0.95\% \\ 
%  625 & 75\% & 75.71\% $\pm$ 1.20\% \\ 
%  625 & 90\% & 58.60\% $\pm$ 1.31\% \\ 
%  \hline
% \end{tabular}
% \end{center}
% \end{table}

% \begin{center}
% \begin{figure}
%   \centering
%   \includegraphics[width=0.5\textwidth]{sparse_weights.png}
%   \caption{Network weights - 90\% sparsity}
%   \label{fig:sparse_filters}
% \end{figure}
% \end{center}

%-------------------------------------------------------------

\section{Conclusions and Future Work}
\label{sec:conclusions}

We have demonstrated a learning scheme with spiking neural networks with self-organization, classification, and clustering properties in an unsupervised fashion. We have found that, in using our training algorithm, the LM-SNN tends to cluster the categories of the inputs into groups of similar filter representations. Moreover, the two-level inhibition scheme tends to create filter maps which vary smoothly between input classes, but our classification scheme does not yet exploit this clustering to improve test accuracy. Nevertheless, we discover that tuning the $c_{\textrm{inhib}}$ parameter dynamically during training allows control over the smoothness of the learned filter map. We also demonstrated the importance of spike ordering for classification by using the \textit{n-gram} scheme. The development of neuron labeling and subsequent classification strategies will continue to be an important part of evaluating SNN learned representations.

Similar to SOM behavior, when an input is presented to the excitatory layer, neurons \textit{compete} to fire. When a neuron fires, it inhibits all other neurons beside its neighbors (the level and pattern of which depends on the choice of inhibition strategy), which encourages neighboring neurons near their threshold to spike. As a result of these spikes, filter weights get closer to the current input and groups of similar filters are formed spontaneously (see Figure \ref{fig:Learned filters}, compared to Figure \ref{fig:Two-level figure}). By relaxing the scheme of large constant inhibition, we allow multiple neurons to fire at once, enabling faster learning and sparse network activation.

Due to the robustness of the LM-SNN in the event of missing inputs, networks can be trained and evaluated on unreliable hardware components. That is, we observe a \textit{graceful degradation} in performance as we remove synapses learned using STDP. The relative infrequency of spikes in both the input and excitatory layer means that the number of STDP updates is low enough to train large LM-SNNs. This is contrast to deep learning neural networks, in which all network parameters are modified synchronously at each back-propagation pass. The asynchronous nature of our system enables this computational advantage.

In this paper we demonstrate a change in SNN training that is accomplished without back-propagation and without labeled data, comprising a system which is suited to rapid decision-making. Once the learning procedure has converged, one may choose to use an auxiliary labeled dataset and network activity together to determine \textit{what} the excitatory neurons represent. Or, one can instead cluster or organize the learned representation as a means of visualizing or summarizing a dataset. Our system represents a step forward in unsupervised learning with spiking neural networks, and suggests additional, biologically inspired research directions. Additional inhibition schemes may lead to a more biologically plausible and useful distribution of network activity.

\section*{Acknowledgements}

This work has been supported in part by Defense Advanced Research Project Agency Grant, DARPA/MTO HR0011-16-l-0006 and by National Science Foundation Grant NSF-CRCNS-DMS-13-11165.

\bibliographystyle{IEEEtran}
\bibliography{Dan_Zotero.bib,additional.bib}

\end{document}